\documentclass{article}
\usepackage{ijcai19}

\usepackage{times}
\usepackage{soul}
\usepackage{url}
\usepackage[hidelinks]{hyperref}
\usepackage[utf8]{inputenc}
\usepackage[small]{caption}
\usepackage{graphicx}
\usepackage{amsmath}
\usepackage{booktabs}
\usepackage{verbatim}
\usepackage{bm}
\usepackage{algorithm}
\usepackage[noend]{algpseudocode}
\usepackage{nicefrac}
\usepackage{xcolor}

\usepackage{float}
\usepackage{todonotes}

\title{Random Sum-Product Forests with Residual Links}

\author{
Fabrizio Ventola$^{1,2}$
\and
Karl Stelzner$^2$\and
Alejandro Molina$^2$\And
Kristian Kersting$^{2,3}$\\
\affiliations
$^1$Research Training Group AIPHES, TU Darmstadt, Germany\\
$^2$Machine Learning Group, Computer Science Department, TU Darmstadt, Germany\\
$^3$Centre for Cognitive Science, TU Darmstadt, Germany\\
\emails
\{ventola@aiphes, stelzner@cs, molina@cs, kersting@cs\}.tu-darmstadt.de
}

\begin{document}

\maketitle

\begin{abstract}
Tractable yet expressive density
estimators are a key building block of probabilistic
machine learning. While sum-product networks (SPNs)
offer attractive inference capabilities, obtaining
structures large enough to fit complex, high-dimensional data
has proven challenging.
In this paper, we present random sum-product forests
(RSPFs), an ensemble approach for mixing multiple
randomly generated SPNs.
We also introduce residual links, which reference specialized substructures of 
other component SPNs in order to leverage
the context-specific knowledge encoded within them.
Our empirical evidence  demonstrates that RSPFs provide better
performance than their individual
components. Adding residual links improves the models further,
allowing the resulting ResSPNs to be competitive with commonly
used structure learning methods.
\end{abstract}

\section{Introduction}
Arguably the most general approach to unsupervised machine learning is
\emph{density estimation}, whereby the goal is to learn the joint probability
distribution underlying a given dataset. Given an estimated distribution,
it can, in principle, be used to solve arbitrary classification or regression tasks via
probabilistic inference, or even to generate new data via sampling.
Traditionally, joint probability distributions have been specified compactly as probabilistic graphical models (PGMs) 
using conditional independences between random variables (RVs).
Recently, a number of deep generative models have also been proposed such as 
variational autoencoders \cite{Rezende2014,Kingma2013},
generative adversarial networks \cite{Goodfellow2014},
normalizing flows \cite{dinh2016density,kingma2018glow},
and neural autoregressive models \cite{Larochelle2011,vandenOord2016,van2016conditional}.
While these models achieve high expressivity by leveraging the representational power of deep neural
networks, they are also highly intractable and generally rely on approximate inference methods such as
MCMC or variational inference \cite{pmlr-v70-hoffman17a,ranganath2014black}.

One promising approach for gaining representational efficiency while remaining tractable
is to directly leverage \emph{context-specific independencies} in a computational graph
for joint probabilities. This is the idea underlying \emph{sum-product networks} (SPNs), 
a deep but tractable family of density estimators~\cite{Poon2011}. In their basic
form, SPNs represent distributions as a deep network where the ``neurons'' are either mixtures (sums), context-specific
independencies (products), or primitive input distributions. They are tractable in the sense that
any marginal or conditional probability can be computed exactly in time linear in the network size.

While the parameters of a given SPN can be readily optimized via stochastic gradient descent
(SGD) or expectation maximization (EM), obtaining a suitable structure is much more difficult.
Structure learning algorithms such as LearnSPN \cite{Gens2013} typically rely on independence tests, which are hard
to scale to large networks and datasets \cite{NIPS2007_3164}. Random SPN structures~\cite{Peharz2019} on the other
hand run the risk of making wrong independence assumptions, leading to suboptimal performance, although they have been proven beneficial in challenging computer vision tasks~\cite{stelzner2019icml_SuPAIR}. 

To address these issues, we explore the combination of random SPN structures and ensemble methods.
Specifically, inspired by random forests~\cite{Breiman2001},
we introduce Random Sum-Product Forests (RSPFs) ---an ensemble approach mixing a set of 
Extremely Random SPNs (ExtraSPNs).
Additionally, in order to make use of the highly specialized substructures of learned SPNs,
we propose to include \emph{residual links}, i.e. references to nodes in other SPNs.
The resulting Residual Sum-Product Networks (ResSPNs) may be thought of as a probabilistic analog
to ResNets~\cite{HeZRS16}, as they rely on refining density estimates made by simple components. 
Our experimental evidence on a variety of datasets demonstrates that SPNs indeed improve
in performance when combined in this manner.

We proceed as follows. We start off by briefly reviewing SPNs. We then introduce random sum-product forests, extremely random SPNs, and residual SPNs. Before concluding, we present our experimental evaluation.

\section{Sum-Product Networks (SPNs)}
Sum-Product Networks are a family of tractable deep density estimators first presented
in \cite{Poon2011}. They have been successfully applied on a variety of domains such
as computer vision \cite{Amer2015}, \cite{Yuan2016}, natural language processing \cite{Cheng2014} \cite{Molina2017}, 
speech recognition \cite{Peharz2014a} and bioinformatics \cite{Ratajczak2014}.
In this section, we first present a formal definition of SPNs, and then introduce commonly used
inference and structure learning methods.

\subsection{Definition of SPNs}
\begin{figure}[tbp]
\begin{center}
  \includegraphics[width=0.8\linewidth]{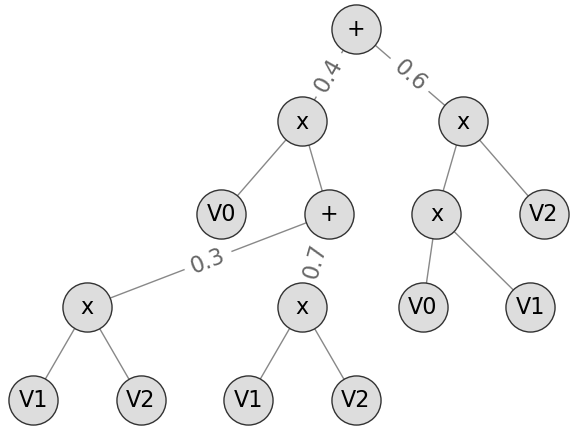}
\end{center}  
  \caption{An SPN over three random variables V0, V1, and V2.  \label{fig:basicspn}}
\end{figure}

An SPN $S$ is a computational graph defined by a rooted DAG, encoding a
probability distribution $P_{\bm{X}}$ over a set of RVs $\bm{X} =
\{X_1, \ldots, X_n\}$, where inner nodes can be either sum or product
nodes over their children (graphically denoted respectively as $\bigoplus$ and $\bigotimes$), and leaves are
univariate distributions defined on one of the RVs $X_{i} \in \bm{X}$.
Each node $n \in S$ has a \emph{scope} $\textsf{scope}(n) \subseteq \mathbf X$,
 defined as the set of RVs appearing in its descendant leaves.  
The subnetwork $S_i$, rooted at node $i$, encodes a distribution over
its scope i.e. $S_{i}(\bm{x}) = P_{\bm{X}_{|scope(i)}}(\bm{x})$ for each $\bm{x} \sim \bm{X}_{|scope(i)}$.
Each edge $(i, j)$ emanating from a sum node $i$ to one of its children $j$
 has a non-negative weight $w_{ij}$, with $\sum_j w_{ij} = 1$.
Sum nodes represent a mixture over the probability
distributions encoded by their children,
while product nodes identify factorizations over contextually independent distributions.
In summary, an SPN can be viewed as a deep hierarchical mixture model of different
factorizations, where the hierarchy is based on the scope of the nodes w.r.t. the
whole set of RVs $\bm{X}$.
An illustration of an SPN defined over three RVs is given in Fig.~\ref{fig:basicspn}.

In order to encode a valid probability distribution, an SPN has to fulfill two structural
requirements~\cite{Poon2011}. First, the scopes of the children of each product
node need to be disjoint (\emph{decomposability}). Second, the scopes of the
children of each sum node need to be identical (\emph{completeness}).
In a valid SPN, the probability assigned to a given state $\bm{x}$ of the RVs $\bm{X}$
can be read out at the root node, and will be denoted $S(\bm{x}) = P_{\bm{X}}(\bm{X} = \bm{x})$.

\subsection{Probabilistic Inference within SPNs}

Given an SPN $S$,  $S(\bm{x})$ can be computed by evaluating the network bottom-up. 
When evaluating a leaf node $i$, $S_i(\mathbf x_{|\textsf{scope}(i)})$ corresponds
to the probability of the state $\mathbf x_{|\textsf{scope}(i)}$ for the RV $\in \textsf{scope}(i)$: $S_i(\mathbf x) = P_{\bm{X}_{|scope(i)}}(\bm{X}_{|\textsf{scope}(i)} = \mathbf x_{|\textsf{scope}(i)})$. 
The value of a product node corresponds to
the product of its children values:
$S_{i}(\bm{x}_{|\textsf{scope}(i)})=\prod_{i \rightarrow j \in S} S_{j}(\bm{x}_{|\textsf{scope}(j)})$;
while, for a sum node its value corresponds to the weighted sum of its children values: 
$S_i(\bm x_{|\textsf{scope}(i)}) = \sum_{i \rightarrow j \in S}w_{ij} S_{j}(\bm{x}_{|\textsf{scope}(j)})$.

All the marginal probabilities,
the partition function, and even approximate MPE queries and states can be computed in time
linear in the \emph{size} of the network i.e. its number of edges as shown
in~\cite{Gens2013,Peharz2015b}.

\subsection{Structure Learning of SPNs}
The prototypical structure learning algorithm for SPNs is LearnSPN~\cite{Gens2013}.
It provides a simple greedy learning schema to
infer both the structure and the parameters of an SPN from data. To this end, 
LearnSPN executes a greedy top-down structure search in the space of \emph{tree-shaped} SPNs, i.e.,
networks in which each node has at most one parent, as summarized in Alg.~\ref{algo:learnspn}.

\begin{algorithm}[t]
  \caption{\textsf{LearnSPN}($\mathcal{D}$, $\bm{X}$, $\mu$, $\rho$)
  \label{algo:learnspn}}
  \begin{algorithmic}[1]
    \Require a set of samples $\mathcal{D}$ over RVs $\bm{X}$; 
    $\mu$: minimum
    number of instances to split; 
    $\rho$: statistical independence threshold
    \Ensure an SPN $S$ encoding a pdf over $\bm{X}$ learned from $\mathcal{D}$
    \If {$|\bm{X}| = 1$}
    \State $S\leftarrow \mathsf{univariateDistribution}(\mathcal{D}, \bm{X})$
    \ElsIf {$|\mathcal{D}| < \mu$}
    \State $S \leftarrow \mathsf{fullFactorization}(\mathcal{D}, \bm{X})$
    \Else
    \State $\{\bm{X}_{d_{1}},\bm{X}_{d_{2}}\}\leftarrow \mathsf{splitFeatures}(\mathcal{D}, \bm{X}, \rho)$
    \If {$|\bm{X}_{d_{2}}| > 0$}
    \State $S\leftarrow \prod_{j=1}^2\mathsf{LearnSPN}(\mathcal{D}, \bm{X}_{d_{j}}, \mu, \rho)$
    \Else
    \State $\{\mathcal{D}_{i}\}_{i=1}^R\leftarrow \mathsf{clusterInstances}(\mathcal{D}, \bm{X})$
    \State $S\leftarrow \sum_{i=1}^R\frac{|\mathcal{D}_i|}{|\mathcal{D}|}\mathsf{LearnSPN}(\mathcal{D}_{i}, \bm{X}, \mu, \rho)$
    \EndIf
    \EndIf
    \Return $S$
  \end{algorithmic}
\end{algorithm}

Specifically, 
LearnSPN recursively partitions the training data matrix $\mathcal{D}$
consisting of i.i.d instances over $\bm{X}$, the set
of columns, i.e., the RVs. 
For each call on a data slice, LearnSPN first tries
to split the data slice by columns.
This is done by splitting the current set of RVs into different groups
such that the RVs in a group are statistically dependent while the
groups are independent, i.e., the joint distribution factorizes over
the groups of RVs. We denote this procedure as \textsf{splitFeatures}.
If no independencies among features/RVs are found, i.e. the splitting fails,
LearnSPN tries to cluster similar data slice rows
(procedure \textsf{clusterInstances}) into groups of similar instances.
In the original work of~\cite{Gens2013}, an online version of hard
Expectation-Maximization (EM) algorithm is employed for this clustering step.
Depending on the assumptions on the distribution of
$\mathbf{X}$, other clustering and splitting algorithms may be more
suitable and may be employed instead~\cite{Vergari2015,Jaini2016,Molina2017,Molina2018}.
When a column split succeeds, LearnSPN adds a product node to the network
whose children correspond to partitioned data slices.
Similarly, after a row clustering step, it adds a sum node where children weights represent
the proportions of instances falling into the obtained clusters.
LearnSPN stops in two cases: (1) when the current data slice contains only one
column or (2) when the number of its rows falls under a certain threshold $\mu$. 
In the first case, a leaf node, representing a univariate distribution, is
introduced by a maximum likelihood estimation from the data slice. 
In the second case, the data slice's RVs are modeled as a full factorization:
they are assumed to be independent and a product node
is put over a set of univariate leaf nodes (each of them estimated as described
for the first case).

Indeed, clustering and splitting
influence each other in terms of quality \cite{Vergari2015}. If a good instance clustering is achieved then
it is likely to enhance the variable splitting and the other way around. This also holds for more advanced structure learning approach beyond LearnSPN, too. 
For instance, \citeauthor{Peharz2013}~\shortcite{Peharz2013} introduced a bottom-up approach to learn SPN structures, using an information-bottleneck method. \citeauthor{Vergari2015}~\shortcite{Vergari2015} employed multivariate leaves for regularization. \citeauthor{Adel2015}~\shortcite{Adel2015} made use of efficient SVD-approaches. \citeauthor{Rahman2016}~\shortcite{Rahman2016} compressed tree-shaped structures into general DAGs. \citeauthor{jaini2018}~\shortcite{jaini2018} estimated product nodes via multi-view clustering over variables.
\citeauthor{DiMauro18}~\shortcite{DiMauro18} investigated approximate independence testing,
and \citeauthor{Molina2018}~\shortcite{Molina2018} non-parametric independence tests for 
learning SPN structures over hybrid domains.
\citeauthor{Rooshenas2014}~\shortcite{Rooshenas2014} refined LearnSPN by learning leaf distributions using Markov networks represented by arithmetic circuits \cite{Lowd2013}.
The resulting SPN learner, called ID-SPN, is state-of-the-art in density estimation on binary data, when considering singleton models. Of course, any of the structure learning approaches can be improved by ensemble and boosting methods~\cite{Liang2017,diMauro2017,ramanan2019tpm}.

\begin{algorithm}[t]
  \caption{\textsf{RSPF}($\mathcal{D}$, $\bm{S}$)
  \label{algo:rspf}}
  \begin{algorithmic}[1]
    \Require a set of samples $\mathcal{D}$ over RVs $\bm{X}$; 
    a set of input SPNs to be combined $\bm{S} = \mathcal{S}_{1}, \dots, \mathcal{S}_{n}$
    \Ensure a $\mathit{RSPF}$ refining the input SPNs $\bm{S}$ on $\mathcal{D}$
    \State $\mathit{RSPF}\leftarrow \sum_{i=1}^n \mathcal{S}_i/n$
    \State $RSPF \leftarrow \mathsf{optimize}(\mathit{RSPF}, \mathcal{D})$\\
    \Return $\mathit{RSPF}$
  \end{algorithmic}
\end{algorithm}

\section{Random Sum-Product Forests}
While SPNs have attractive inference properties, scaling them in a manner similar to deep
neural networks has proven challenging.
The structural constraints which SPNs have to obey in order to guarantee validity are a major
reason for this.
They necessitate the careful design of structures, either by hand or through learning from data.
Unfortunately, learning SPN structures (as sketched by LearnSPN)
has proven hard to scale. This is mainly caused by the cost of determining how to split
variables: Ideally, one would always determine the two subsets with minimum empirical
mutual information.
This takes cubic time, however~\cite{Gens2013} and is thus much too slow.
Therefore, one typically considers only pairwise dependencies in practice, reducing the cost
to quadratic time, which is unfortunately still prohibitive for many applications.
Recently, \citeauthor{Peharz2019}~\shortcite{Peharz2019} proposed random tensorized SPNs
(RAT-SPNs), which forgo structure learning in favor of randomly generated structures.

\begin{algorithm}[t]
  \caption{\textsf{ExtraSPN}($\mathcal{D}$, $\bm{X}$, $\mu$, $\beta$)
  \label{algo:extraspn}}
  \begin{algorithmic}[1]
    \Require a set of samples $\mathcal{D}$ over RVs $\bm{X}$; 
    $\mu$: minimum
    number of instances to split; 
    $\beta$: how likely the random feature splitting or random clustering could fail 
    \Ensure an SPN $S$ encoding a pdf over $\bm{X}$ learned from $\mathcal{D}$
    \If {$|\bm{X}| = 1$}
    \State $S\leftarrow \mathsf{univariateDistribution}(\mathcal{D}, \bm{X})$
    \ElsIf {$|\mathcal{D}| < \mu$}
    \State $S \leftarrow \mathsf{fullFactorization}(\mathcal{D}, \bm{X})$
    \Else
    \State $\{\bm{X}_{d_{1}},\bm{X}_{d_{2}}\}\leftarrow \mathsf{randomSplitFeatures}(\bm{X}, \beta)$
    \If {$|\bm{X}_{d_{2}}| > 0$}
    \State $S\leftarrow \prod_{j=1}^2\mathsf{ExtraSPN}(\mathcal{D}, \bm{X}_{d_{j}}, \mu, \beta)$
    \Else
    \State $\{\mathcal{D}_{i}\}_{i=1}^R\leftarrow \mathsf{clusterInstances}(\mathcal{D}, \bm{X})$ //or random
    \State $S\leftarrow \sum_{i=1}^R\frac{|\mathcal{D}_i|}{|\mathcal{D}|}\mathsf{ExtraSPN}(\mathcal{D}_{i}, \bm{X}, \mu, \beta)$
    \EndIf
    \EndIf
    \Return $S$
  \end{algorithmic}
\end{algorithm}

While their approach of scaling random SPN structures yielded surprisingly good results,
it carries the risk of introducing false independency assumptions.
To repair such mistakes, we first propose an ensemble approach similar to Random Forests~\cite{Breiman2001},
called Random Sum-Product Forests (RSPFs). As summarized in Alg.~\ref{algo:rspf},
it generates a number of individual SPNs which are then mixed together by adding a single sum node
on top of the individual SPNs.
This way, all mixed SPNs can be trained jointly using stochastic gradient or EM. 

In principle, any SPN learned via structure learning (potentially in combination with bagging)
or just RAT-SPNs can be used as components in this ensemble.
Here, we opt for a faster and even simpler method: 
Akin to extremely randomized trees~\cite{geurtsEW06},
each individual SPN is trained using the whole learning sample 
and the top-down splitting in LearnSPN is randomized, as summarized in Alg.~\ref{algo:extraspn}.
That is, instead of computing the locally optimal splitting of random variables based on, e.g., mutual information or independence tests, a random split is selected. The resulting \emph{extremely randomized SPNs} (ExtraSPNs) are similar in spirit to extremely randomized cutset networks~\cite{diMauro2017}, except that ExtraSPNs do not use any statistical test for
the splitting procedure. Instead, we fix a parameter $\beta$ to set the probability
of the splitting test to fail, and randomly group the features in two clusters.
In this way, the splitting procedure has constant complexity instead of the quadratic
complexity of the G-test employed in the original LearnSPN. 
Moreover, to foster diversity in the set of ExtraSPNs, we sample the
parameter $\mu$ at random from the range $[1, |\mathcal{D}|/\gamma]$ for each SPN.
Here, $\mathcal{D}$ is the dataset and $\gamma$ is a hyperparameter that can shrink the
range of the possible values of $\mu$. When $\gamma$ is not set to $1$ then the range
of the possible values for $\mu$ is smaller, in this way we can both tune the learning
times of the ExtraSPNs and also control their depth.

\begin{figure}[t]
\begin{center}
  \includegraphics[width=0.8\linewidth]{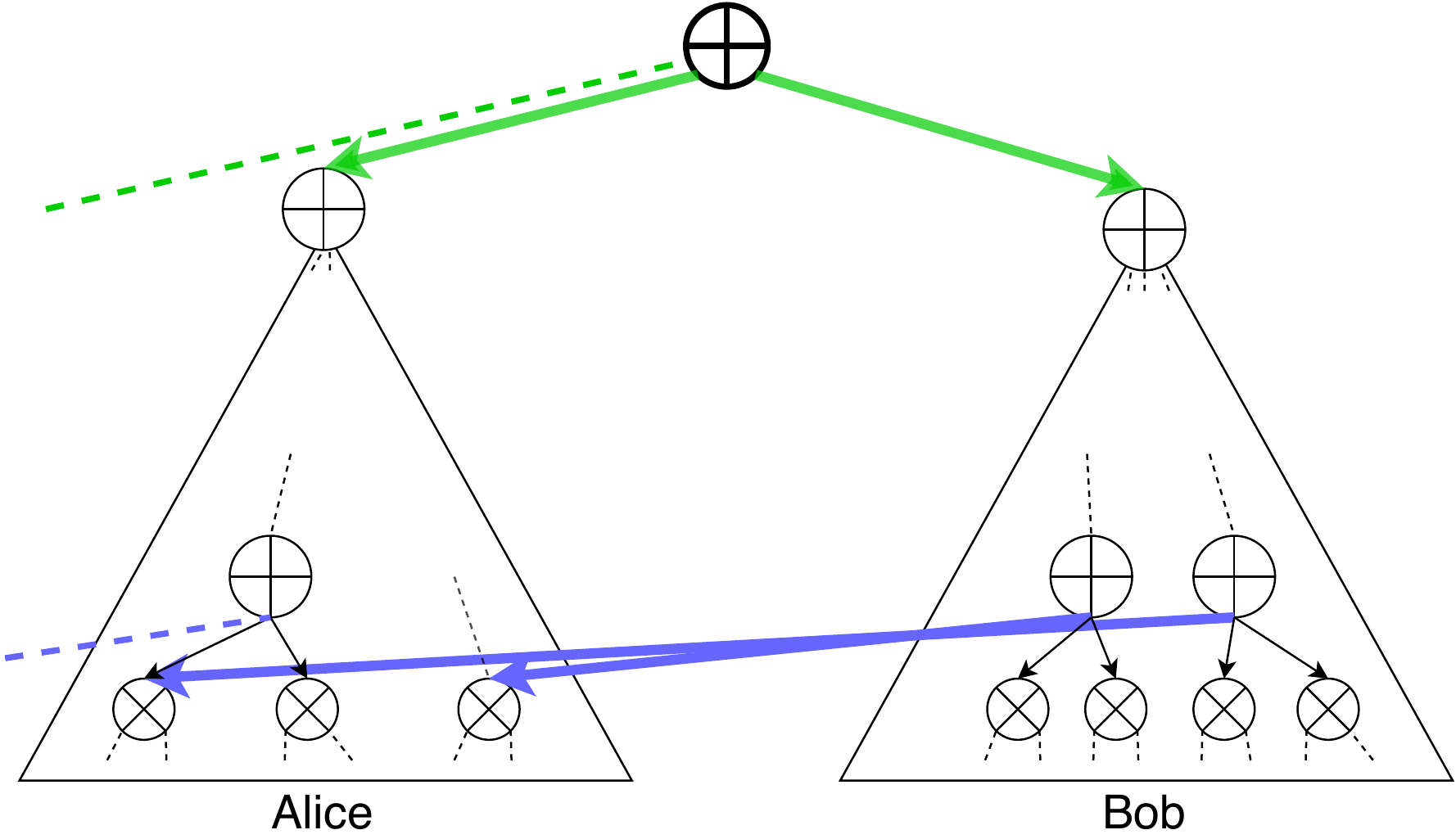}
  \end{center}
  \caption{Illustration of Residual Sum-Product Learning. The blue edges denote residual links between the right (Bob) and left (Alice) SPN, which let Bob directly fit a residual.
  Green edges realize the global mixture to form a Sum-Product Random Forest. \label{fig:resspn}}
\end{figure}

\section{Residual Sum-Product Learning}
Random Sum-Product Forests suggest that learning ensembles of SPNs is as easy as mixing
adding a single sum node at the top.
Here, we aim to improve on this by making use of the fact that SPNs consist of highly
specialized local substructures. This suggests that an ensemble method for SPNs
should also work at the level of these substructures.
Instead of hoping that each SPN fits a desired contribution to the global joint distribution,
and inspired by ResNets~\cite{HeZRS16} and BoostResNet~\cite{Huang18}, we explicitly
let them fit residuals.
Consider two SPNs called Alice and Bob as shown in Fig.~\ref{fig:resspn}.
Our goal is to improve the performance of one of Bob's sum nodes
using a node $a(\bm{z})$ of the same scope from Alice.
We denote Bob's desired density as $b(\bm{z})$.
Instead of learning this density directly, we instead let Bob fit a residual, namely,
$r(\bm{z}) = b(\bm{z})-a(\bm{z})$, up to proper weighting.
That is, Bob's original density is recast into $b(\bm{z}) = a(\bm{z})+ r(\bm{z})$,
up to proper weighting. This can be achieved in a simple way: We add Alice's product
node as a child to Bob's sum node, cf.~\ref{fig:resspn}.
Adding such residual links between individual SPNs does not affect the
validity of individual SPNs, since the scope of the
two nodes is the same --- the sum-product networks remain complete.
For additional flexibility, we can also pick one of Alice's nodes $a$ with
\textsf{scope}($b$) $\subset$ \textsf{scope}($a$), and marginalize the surplus
random variables from $a$ before adding the connection.
The weights of the new mixture components can be set proportionally
to their training data slices or eventually  
optimized jointly in a supplementary parameter estimation step,
even after mixing Alice and Bob additionally using a single sum node at the top,
such as in RSPFs.  
Thus, we propose to apply residual sum-product learning to every ExtraSPN
within a RSPF. To do so, we generate an array of ExtraSPNs
$E_1,E_2,\ldots, E_n$, pick a random one $E_i$, and introduce residual links between each
pair $(E_i,E_{j})$. We add a top sum node to form a RSPF
and finally train all parameters jointly.
We name the resulting model Residual SPNs (ResSPNs).

Overall, we hypothesize that it is easier to optimize the residual density than to optimize
the original, unreferenced density. To the extreme, if a density from Alice was optimal, 
it would be easier to push the residual in Bob to zero than to fit the density
in Bob from scratch or by mixing Alice and Bob at the top node only.
As a result, we build a stronger density estimator
by referencing to local \emph{context-specific} substructures.
In fact, residual connections make
SPNs wider as also suggested by Fig.~\ref{fig:resspn} and, in turn,
increase the variance in sizes of training data slices used to train mixture components.
This is akin to what is known 
for residual neural networks~\cite{veitWB16} but now for hierarchical mixture models. 

Different procedures are imaginable for selecting the pairs of nodes between
which edges are to be added.
For this reason, one can see the Residual
Sum-Product Learning as a general schema, where different strategies for 
adding residual links may be selected based on the problem at hand.
Here, we adopt a randomized greedy approach as summarized in Alg.~\ref{algo:resspn}.
From the set of input SPNs, we randomly select one to which all future edges will be added.
Then, we iteratively add references to each of the remaining input SPNs $S$, where the
maximum percentage of nodes from which to draw connections to each one is given
by the hyperparameter $k$. The connections are drawn in a greedy fashion: We start by
selecting candidate nodes from our ResSPN by traversing it via breadth-first-search (BFS),
excluding the root. For each visited node, we try to find a suitable partner in $S$,
by also traversing it via BFS. If we find a suitable pairing, i.e. one where the scopes
match as described above, we draw a residual link. This continues until either all pairs of
nodes have been considered or the maximum number of connections has been drawn.
Alternatively, one may also consider an \emph{informed}
variant where residual links are added only if they locally improve the learning objective
on the training or a validation set. We name this variant InfoResSPN.

\begin{algorithm}[t]
  \caption{\textsf{ResSPN}($\mathcal{D}$, $\bm{S}$, $k$)
  \label{algo:resspn}}
  \begin{algorithmic}[1]
    \Require a set of samples $\mathcal{D}$;
    a set of input SPNs to be combined $\bm{S} = \mathcal{S}_{1}, \dots, \mathcal{S}_{n}$;
    the ratio of nodes from which to draw residual links $k$
    \Ensure a $\mathit{ResSPN}$ built from $\bm{S}$ and optimized on $\mathcal{D}$
    \State $\mathit{ResSPN} \leftarrow \textsf{copyRandomElement}(\bm{S})$
    \For{$S \in \bm{S} \setminus \mathit{ResSPN}$}
        \State $\eta = 0$
        \For {$s_1 \in$ \textsf{BFS}($\mathit{ResSPN}$)}
            \For {$s_2 \in$ \textsf{BFS}($S$)}
                \If{\textsf{scope}($s_1$) $\subseteq$ \textsf{scope}($s_2$)}
                    \State $s_2' \leftarrow $\textsf{Marginalize}($s_2$, \textsf{scope}$(s_1)$)
                    \State Add connection from $s_1$ to $s_2'$
                    \State $\eta \leftarrow \eta + 1$
                \EndIf
            \EndFor
            \If{$\eta >k |\mathit{ResSPN}|$}
                \State \textbf{break}
            \EndIf
        \EndFor
    \EndFor
    \State $\mathit{ResSPN} \leftarrow \mathsf{RSPF}(\mathcal{D}, \bm{S} \cup \{\mathit{ResSPN}\})$\\
    \Return $\mathit{ResSPN}$
  \end{algorithmic}
\end{algorithm}

\section{Experimental Evaluation}

With our experimental evaluation, we aim to investigate whether the presented ensemble
methods provide the intended benefits. In the process, we also explore how SPNs
benefit from wider and deeper structures.
Specifically, we examine the following questions:
\begin{description}
    \item[\textbf{(Q1)}] Can RSPFs improve upon singleton SPNs learned by LearnSPN?
    \item[\textbf{(Q2)}] Can RSPFs repair wrong independencies?
    \item[\textbf{(Q3)}] Can residual links improve the performance of RSPFs?
    \item[\textbf{(Q4)}] Can informed residual links improve compared to non-informed ones?
\end{description}
To answer these questions, we learned a RSPF with and without residual links on standard benchmarks
summarized in Tab.~\ref{tab:datasets}, split into training, validation and test sets.
For parameter estimation we used the EM algorithm. The datasets are a subset of the ones used e.g. in \cite{Gens2013} and \cite{Rooshenas2014}.
They were firstly introduced in \cite{Lowd2010} and in \cite{Haaren2012}. All of them are
binary datasets and most of them are the binarized version of some popular datasets of the UCI machine learning repository \cite{UCI98}.
All algorithms were implemented
in Python\footnote{Code available at: \url{https://github.com/ml-research/resspn}} making use of the \emph{SPFlow} library~\cite{SPFlowLib}. All experiments were run on a DGX-2 system.

\newcommand*\mccol[2]{\multicolumn{#1}{c}{#2}}
\newcommand*\tmccol[2]{\mccol{#1}{\tiny\textsf{#2}}}
\newcommand*\bmccol[2]{\mccol{#1}{\textbf{#2}}}

\begin{table}[!tbp]
  \centering
    \scriptsize

  \setlength{\tabcolsep}{5pt}
  \begin{tabular}{r r r r r r}
  \toprule
    & $|\bm{X}|$ & $|T_{train}|$ & $|T_{val}|$ & $|T_{test}|$  \\
    \midrule
    \textbf{NLTCS}        & 16             & 16181         & 2157        & 3236                  \\ 
    \textbf{MSNBC}        & 17             & 291326        & 38843       & 58265               \\
    \textbf{Plants}       & 69             & 17412         & 2321        & 3482                  \\
    \textbf{Audio}        & 100            & 15000         & 2000        & 3000                   \\
    \textbf{Jester}       & 100            & 9000          & 1000        & 4116                    \\
    \textbf{Netflix}      & 100            & 15000         & 2000        & 3000                   \\
    \bottomrule
  \end{tabular}
   \caption[datasets]{Statistics of the datasets used. \label{tab:datasets}}
\end{table}

\begin{table}[t]
  \centering
  \scriptsize
  \setlength{\tabcolsep}{5pt}
 
  \begin{tabular}{r r r r | r r}
    \toprule
     & \bmccol{5}{Test Accuracy}  \\
    \midrule
     & \bmccol{3}{Baseline Singleton SPN Learners} & \bmccol{2}{Ensemble SPN}  \\
     & \tmccol{1}{Best ExtraSPN} & \tmccol{1}{LearnSPN} & \tmccol{1}{ID-SPN} & \tmccol{1}{RSPF} & \tmccol{1}{ResSPN}  \\
    \midrule 
    \textbf{NLTCS}      & -6.153 & -6.11 & {\bf -6.02}  & -6.046$\Uparrow$ & -6.040$\bullet$  \\
    \textbf{MSNBC}     & -6.433 & -6.11 & {\bf -6.04}  & -6.104$\Uparrow$ & -6.097$\bullet$  \\
    \textbf{Plants}     & -16.683 & -12.98 & {\bf -12.54} & -14.573$\uparrow$ & -13.908$\bullet$   \\
    \textbf{Audio}      & -42.639 & -40.5 & {\bf -39.79}  & -40.833$\uparrow$ & -40.762$\bullet$    \\
    \textbf{Jester}     & -55.335 & -53.48 & {\bf -52.86}  & -53.885$\uparrow$ & -53.995$\uparrow$   \\
    \textbf{Netflix}    & -60.330 & -57.33 & {\bf -56.36}  & -57.900$\uparrow$ & -57.867$\bullet$   \\
    \bottomrule 
   \end{tabular}
   \caption[Accuracy]{Average test log-likelihood of RSPF and ResSPN compared to baseline singleton SPN learners.
   The best results including the baselines are shown in bold.
   As one can see, ID-SPN gives the best results, however, it uses Markov networks
   represented as ACs in the leaves.
   This can also be adapted for RSPFs.
   More importantly, ensembles of SPNs improve upon ExtraSPNs, as
   denoted by ``$\uparrow$'', getting sometimes even better
   than LearnSPN, denoted as ``$\Uparrow$''.  Residual links can, in turn,
   improve the performance of RSPFs, as denoted by ``$\bullet$''.
  \label{tab:exps_accuracy}}
\end{table}

\begin{table}[t]
  \centering
  \scriptsize
  \setlength{\tabcolsep}{5pt}
  
  \begin{tabular}{r r r}
    \toprule
     \multicolumn{3}{c}{\textbf{Training Accuracy}}\\
    \midrule
     & \tmccol{1}{RSPF} & \tmccol{1}{ResSPN} \\
    \midrule 
    \textbf{NLTCS}       & -6.006   & -5.995$\bullet$  \\
    \textbf{MSNBC}      &  -6.097  &  -6.090$\bullet$      \\
    \textbf{Plants}     &  -14.056  &  -12.969$\bullet$     \\
    \textbf{Audio}     &  -39.784  &  -39.182$\bullet$   \\
    \textbf{Jester}      &  -52.352  & -52.079$\bullet$    \\
    \textbf{Netflix}     &  -56.586  & -55.982$\bullet$  \\
    \bottomrule 
  \end{tabular}
  \caption[Training Accuracy]{Average training log-likelihood of RSPF and ResSPN.
  The best results are denoted using ``$\bullet$''. ResSPNs consistently achieve a
  better fit of the training data.
  
    \label{tab:train_accuracy}}
\end{table}

\begin{table}[t]
  \centering
  \scriptsize
  \setlength{\tabcolsep}{3pt}
 
  \begin{tabular}{r r r r | r r r}
    \toprule
     & \bmccol{6}{Test Accuracy}  \\
    \midrule
    & \bmccol{3}{With Residual Links} & \bmccol{3}{Without Residual Links}  \\
     & \tmccol{1}{ResSPN 3} & \tmccol{1}{ResSPN 5} & \tmccol{1}{ResSPN 10} & \tmccol{1}{RSPF 3} & \tmccol{1}{RSPF 5} & \tmccol{1}{RSPF 10} \\
    \midrule 
    \textbf{NLTCS}      & -6.118$\bullet$ & -6.068$\bullet$ & -6.040$\bullet$  & -6.192\hspace{1ex} & -6.109\hspace{1ex} & -6.046\hspace{1ex}  \\
    \textbf{MSNBC}     & -6.235$\bullet$ & -6.145$\bullet$ & -6.097$\bullet$  & -6.316\hspace{1ex} & -6.225\hspace{1ex} & -6.104\hspace{1ex} \\
    \textbf{Plants}     & -14.732$\bullet$ & -14.631$\bullet$ & -13.908$\bullet$ & -15.616\hspace{1ex} & -15.161\hspace{1ex} & -14.573\hspace{1ex} \\
    \textbf{Audio}      & -41.492$\bullet$ & -41.433$\bullet$ & -40.762$\bullet$  & -41.883\hspace{1ex} & -41.482\hspace{1ex} & -40.833\hspace{1ex} \\
    \textbf{Jester}     & -53.772$\bullet$ & -54.040\hspace{1ex} & -53.995\hspace{1ex}  & -53.987\hspace{1ex} & -53.734$\bullet$ & -53.885$\bullet$ \\
    \textbf{Netflix}    & -58.609$\bullet$ & -58.509$\bullet$ & -57.867$\bullet$  & -59.121\hspace{1ex} & -58.570\hspace{1ex} & -57.900\hspace{1ex} \\
    \bottomrule 
    \textbf{wins}  & 6/6 & 5/6 & 5/6 & 0/6 & 1/6 & 1/6 
  \end{tabular}
   \caption[Accuracy]{Average test log-likelihood of ResSPN --- RSPF with residual links ---
  for ensembles of 3, 5, and 10 ExtraSPNs compared to RSPFs without residual links. The best results are denoted using ``$\bullet$''.
  Overall, one can clearly see that adding residual links is beneficial.
  The only exception is Jester, where ResSPNs overfit.
  \label{tab:exps_res_vs_ensemble}}
\end{table}

\begin{table}[t]
  \centering
  \scriptsize
  \setlength{\tabcolsep}{5pt}
  
  \begin{tabular}{r r r}
    \toprule
     & \bmccol{2}{Test Accuracy}  \\
    \midrule
      & \tmccol{1}{RSPF w/} & \tmccol{1}{InfoResSPN}\\
      & \tmccol{1}{instance clustering} &   \\
    \midrule 
    \textbf{NLTCS}        &  -6.064  &     -6.020$\bullet$   \\
    \textbf{Audio}       &   -41.919  &    -40.259 $\bullet$ \\
    \textbf{Jester}      &  -53.393  &   -53.214$\bullet$    \\
    \textbf{Netflix}     &   -59.668  &    -58.085$\bullet$  \\
    \bottomrule 
  \end{tabular}
  \caption[Accuracy]{Average test log-likelihood
  of InfoResSPN compared to RSPF based on ExtraSPNs with proper clustering.
   InfoResSPN improves (denoted with ``$\bullet$'') the accuracy of the RSPF baseline.
   The regularization realized by InfoResSPN makes it
   less prone to overfitting (see Jester).
  \label{tab:exps_accuracy_2}}
\end{table}

The following subsections will briefly describe the adopted experimental protocol and discuss
the results for each question {\bf (Q1)}-{\bf (Q4)} in turn. 

\subsection{LearnSPN versus RSPFs -- (Q1)}
To answer \textbf{(Q1)}, we build a RSPF from 10 ExtraSPNs.
For learning the ExtraSPNs, we take the $\mu$ hyperparameter at random in the
range $[1, |D|/5]$ and fix $\beta$ to 0.6. We similarly randomize the instance
clustering step using the same hyperparameter $\beta$.
The RSPF is then built as described in Alg.~\ref{algo:rspf}.
We optimize the resulting structure with EM fixing a limit of 1000 iterations.
The optimization stops if the variance of the training likelihood on the last 5 iterations
is lower than 1e-7. Then, to assess the accuracy of learned estimators, we compute
the average log-likelihood on the test sets. We compare it with the ones obtained
with LearnSPN as reported in \cite{Rooshenas2014}.
Tab.~\ref{tab:exps_accuracy} shows the average test likelihood for LearnSPN and
RSPF. One can see that the ensemble strategy RSPF is competitive
compared to LearnSPN since it can achieve similar accuracy and is more
accurate on NLTCS and MSNBC without involving a meticulous hyperparameter selection. 
Thus, we can answer \textbf{(Q1)} positively: RSPFs improve
upon singleton SPNs learned by LearnSPN.

\subsection{Singleton ExtraSPNs versus RSPFs -- (Q2)}
Here, we consider the same RSPFs as before in the context of \textbf{(Q1)}.
We compare them with the single best available ExtraSPN. To do so, we generate
a set of ExtraSPNs as before, except that we include a proper clustering step
for generating sum nodes as opposed to random clustering. We then optimize
their parameters with EM and report best test log-likelihood achieved among
all of them.
By looking at the results in Tab.~\ref{tab:exps_accuracy}, one can see that the RSPF
always outperforms the best ExtraSPN.
This means that as an ensemble method, RSPF does indeed improve upon its components,
even when they are augmented using proper clustering and optimization.
It also indicates that RSPF is able to
repair false independency assumptions made by the ExtraSPNs.
Therefore, we can affirmatively answer \textbf{(Q2)}.

\subsection{RSPFs with Residual Learning -- (Q2, Q3)}
Regarding these questions, we start from the same ExtaSPNs which
were learned for \textbf{(Q1)} and \textbf{(Q2)}, and use them to build ResSPNs,
according to Alg.~\ref{algo:resspn}. We choose the number $n$ of ensemble components
from $\{3, 5, 10\}$.
After adding the residual links, we optimize the ResSPN structure on
the training set with EM for a maximum number of 1000 iterations.
We stop the optimization earlier if the variance of the training likelihood on
the last 5 iterations is lower than 1e-07. 
We perform a hyperparameter search over $k$  within the range $[0.1,  0.2]$,
and select the models with the best accuracy on a separate validation set.

\begin{figure}[t!]
\centering

  \includegraphics[width=\linewidth,trim=0 62 0 80,clip]{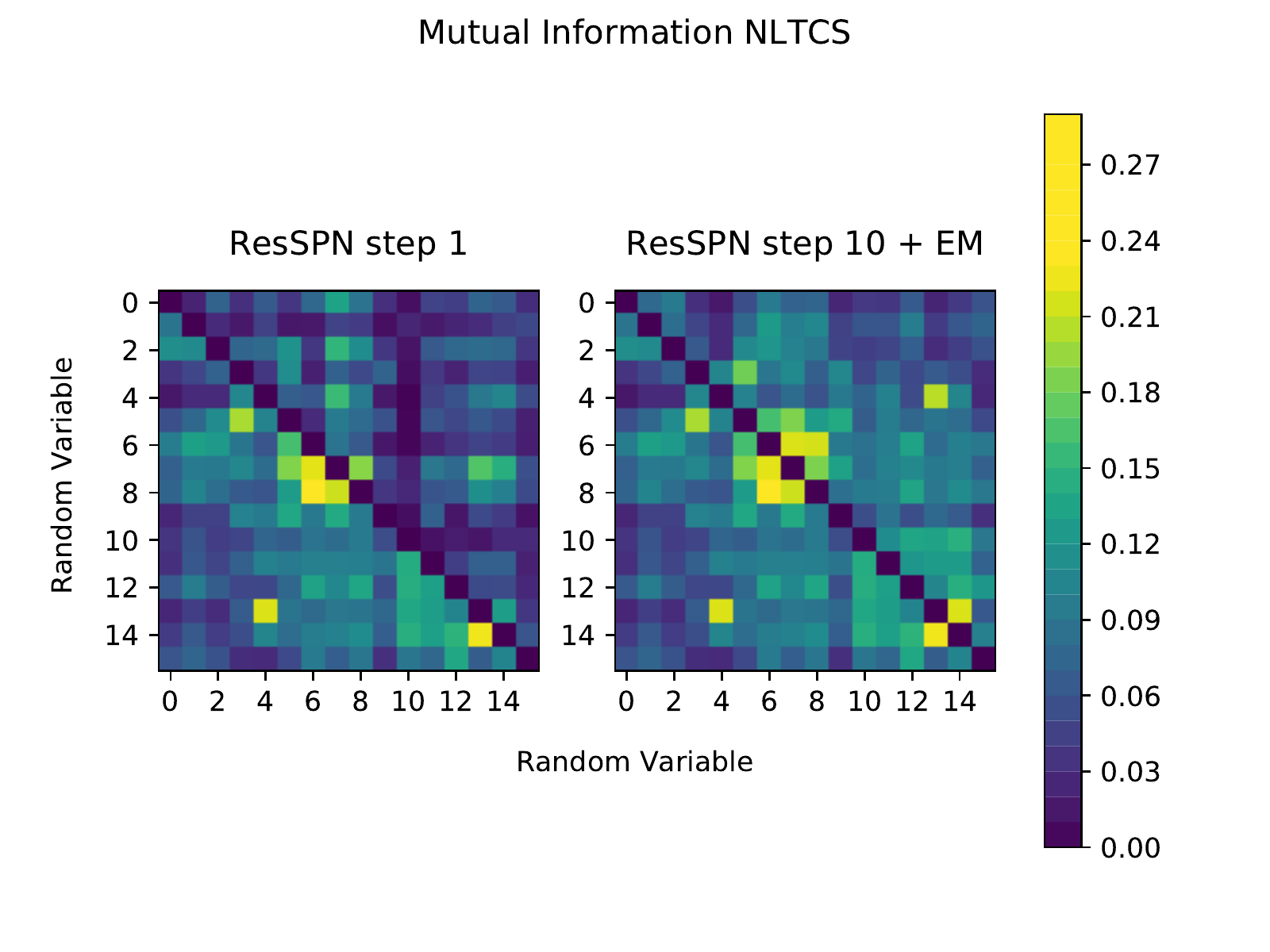}
  
  \caption{
  The lower triangle of the matrices shows the pairwise empirical mutual information
  between variables in the NLCTS dataset. (Upper triangle, left) MI computed
  with a ResSPN after adding links to one other SPN.
  (Upper triangle, right) MI computed with a ResSPN
  after adding links to 9 other SPNs and optimizing with EM for 10 iterations.
  The fact that the MI of the fully trained ResSPN is much closer to the 
  empirical MI indicates that ResSPNs are able to repair wrong independence assumptions.
  \label{fig:mi_nltcs}
  }
\end{figure}

Tab.~\ref{tab:exps_accuracy} summarizes the average test log-likelihood of ResSPNs
and RSPFs. As one can clearly see, residual links generally improve the test accuracy,
with Jester being the only exception.
To investigate the possibility of overfitting, we also analyzed the training likelihood
of both models as reported in Tab.~\ref{tab:train_accuracy}.
One can observe that on all datasets including Jester, the ResSPN indeed achieves
a better training accuracy, indicating that overfitting did indeed occur on Jester.
The dataset appears to be generally prone to overfitting, as even for RSPFs,
better results are achieved with fewer than 10 ensemble components  (Tab.~\ref{tab:exps_res_vs_ensemble}). 
In summary, residual links improve the ability of the model to fit
the training data and they generally also yield a better test accuracy.
These observations also hold when a smaller number of ExtraSPNs are used, 
as can be seen in Tab.~\ref{tab:exps_res_vs_ensemble}.
Given this, we can positively answer \textbf{(Q3)}: residual links indeed improve
the performance of RSPFs.

Additionally, ResSPNs are also capable of repairing wrong independence
assumptions made by the singleton models \textbf{(Q2)}. To demonstrate this,
we compare the empirical pairwise mutual information between random variables on the 
training dataset with the one computed on our model at various stages of training.
Fig.~\ref{fig:mi_nltcs} depicts the empirical pairwise mutual information on the NLTCS dataset
on the lower triangle of the matrices, and the mutual information computed from our model
on the upper triangle. The left matrix uses a ResSPN in the early stages of training, after
the first set of residual links has been added, while the right one shows the result for
a fully trained ResSPN based on 10 ExtraSPNs.
It is apparent that after full training, the ResSPN is much more capable 
of representing the correlations among the random variables, since the
matrix on the right is almost symmetric.

\begin{table}[t]
  \centering
  \scriptsize
  \setlength{\tabcolsep}{5pt}
  
  \begin{tabular}{r r r r r r r}
    \toprule
     & \bmccol{2}{LearnSPN} &  \bmccol{2}{ResSPN 10}  \\
    \midrule
     & \tmccol{1}{edges} & \tmccol{1}{layers} & \tmccol{1}{edges} & \tmccol{1}{layers} \\
    \midrule 
    \textbf{NLTCS}      & 7509   &  4  & 17102  & 169 \\
    \textbf{MSNBC}      &  22350  & 4  & 8040  & 136  \\
    \textbf{Plants}     & 55668   &   6  &  66498 & 298  \\
    \textbf{Audio}      &  70036  &  8  & 153791  & 296  \\
    \textbf{Jester}     &  36528  &  4   &  104053 & 259  \\
    \textbf{Netflix}    &  17742  &  4  & 153791  & 296  \\
    \bottomrule 
  \end{tabular}
  \caption[Structures]{The structure of ResSPNs can be deeper and/or wider compared to 
  singleton tree-SPNs learned via LearnSPN as reported in \cite{Vergari2015}.
    \label{tab:structures}}
\end{table}

\subsection{Regularization via informed residual -- (Q4)}
Finally, we investigate whether using an informed strategy for adding residual links and, hence,
reducing the level of greediness can be beneficial in terms of model accuracy.
To this end, we explore a variant of ResSPNs called InfoResSPN, which takes
advantage of the information provided by input SPNs learned via LearnSPN and its
derivatives. Specifically, we modify line 6 in Alg. \ref{algo:resspn} such that
residual links are only added if they improve the model's
training likelihood on the data slice for which the node $s_1$ is responsible for according
to LearnSPN.

Tab.~\ref{tab:exps_accuracy_2}
summarizes the results, comparing InfoResSPN and RSPF.
In order to get a more challenging baseline, this time RSPFs are composed of
ExtraSPNs learned with regular instance clustering instead of random clustering.
We opt for this stronger baseline
since we want to check whether informed residual links can be successful
at extracting the knowledge contained in each of the ExtraSPNs.
We fix the ratio of candidate nodes for each iteration to $k=0.1$, and use
10 ExtraSPNs on NLTCS and Audio
and 5 for the other datasets. All the models were optimized with EM
for 10 iterations on the training set.

InfoResSPNs improve upon the RSPF baseline.
Moreover, InfoResSPN also achieves overall better accuracy than ResSPN,
and even succeeds on Jester where ResSPN overfitted.
Intuitively, InfoResSPN is less prone to overfitting since its strategy
for adding residual links is stricter compared to ResSPNs.
Thus, we can positively answer to question \textbf{(Q4)}: employing
informed residual links, even if it adds a computational burden, improves the accuracy
of learned models and may help make them less prone to overfitting.\\

\subsection{Discussion of the Experimental Results}
We addressed the questions {\bf (Q1)}-{\bf (Q4)} in order to answer a more general question, namely, whether deeper and wider probabilistic models are actually beneficial. It is well known that learning and scaling 
deep architectures is a challenging problem, and SPNs face this challenge
too.
We already discussed the results in terms of accuracy in the previous sections, thus,
taking into account the structural details of the learned models shown in Tab.~\ref{tab:structures},
we can indeed confirm that wider and deeper SPNs ---learned by adding residual links---
in general, perform considerably better than singleton tree-shaped models.
With ResSPNs, it is possible to obtain 
results competitive to LearnSPN without the need for expensive
hyperparameter selection as well as costly
statistical independence tests and clustering.
In Tab.~\ref{tab:exps_accuracy} one can also see that performance in accuracy is 
comparable to ID-SPN, which is the most accurate method so far while also being
difficult to scale due to the learning of Markov Networks as multivariate leaves 
and the need of an attentive hyperparameter selection.
In contrast, RSPF and ResSPNs naturally scale to thousands of random variables.
In any case, one can, of course, employ ID-SPNs instead of ExtraSPNs
within RSFPs --- they are orthogonal to each other.
Doing so provides an interesting avenue for future work.
This also highlights the peculiarity of ResSPN that can be seen
as a general schema: One can easily extend and plug-in 
different singleton learners, employ different strategies for adding residual links,
and make use of many deep neural learning techniques.

\section{Conclusion}
Deeper and wider probabilistic models are more difficult to train.
To make this task easier, we have
introduced an ensemble learning framework for sum-product networks (SPNs),
which makes it possible to combine arbitrary groups of SPNs into a larger model.
Our  experimental evidence shows that even when the input SPNs are generated
randomly, the resulting random sum-product forests (RSPFs)
are competitive with state of the art SPN learners.
As it turns out, ensembles of random SPNs
can indeed ``repair'' wrong independence assumptions made by individual SPNs.
As an additional improvement to RSPFs, we have proposed
residual links to substructures in other SPNs, yielding ResSPNs.

RSPNs and ResSPNs can be seen as general schemas for building ensembles of SPNs and thus provide
several interesting avenues for future work. One is to explore different structure learners instead of ExtraSPN or even a mix of different ones. 
Furthermore, one should explore other strategies for adding residual links.
For instance, given that we have a tractable model at hand,
one may compute the expected value of adding a residual link.
Post-pruning the final ResSPN, employing dropout as well as online learning
are further interesting avenues. Of course, one should run more experiments,
i.e., on images, text, and hybrid domains.
Most important, however, is to push the merge of deep probabilistic
and neural learning further that \citeauthor{stelzner2019icml_SuPAIR}~\shortcite{stelzner2019icml_SuPAIR} have already shown to be beneficial.

\section*{Acknowledgments}
FV and KK acknowledge the support by the German Research
Foundation (DFG) as part of the Research Training Group
Adaptive Preparation of Information from Heterogeneous
Sources (AIPHES) under grant No. GRK 1994/1. KK also acknowledges the support of the Rhine-Main Universities Network for ``Deep Continuous-Discrete Machine
Learning'' (DeCoDeML).

\bibliographystyle{named}
\bibliography{ijcai19}

\end{document}